%% file: naacl2021.tex
\pdfoutput=1
\documentclass[11pt]{article}

\usepackage[]{naacl2021}

\usepackage{times}
\usepackage{latexsym}

\usepackage[T1]{fontenc}
\usepackage[utf8]{inputenc}

\usepackage{microtype}

\usepackage{url}
\usepackage{hyperref}
\usepackage{graphicx}
\usepackage{caption}
\usepackage{subcaption}
\usepackage{bbm}
\usepackage{algorithm}
\usepackage{listings}
\usepackage{enumerate}
\usepackage{algorithm}
\usepackage[noend]{algpseudocode}
\usepackage{amsmath}
\usepackage{multirow}
\usepackage{tabularx}
\usepackage{array}
\usepackage{scrextend}
\usepackage{booktabs}
\usepackage{algpseudocode}
\usepackage{adjustbox}

\newcommand\Algphase[1]{%
\vspace*{-.4\baselineskip}\Statex\hspace*{\dimexpr-\algorithmicindent-2pt\relax}\rule{\textwidth}{0.4pt}%
\Statex\hspace*{-\algorithmicindent}\textbf{#1}%
\vspace*{-.7\baselineskip}\Statex\hspace*{\dimexpr-\algorithmicindent-2pt\relax}\rule{\textwidth}{0.4pt}%
}

\newcommand\Algphaseonecolumn[1]{%
\vspace*{-.4\baselineskip}\Statex\hspace*{\dimexpr-\algorithmicindent-2pt\relax}\rule{7.5cm}{0.4pt}%
\Statex\hspace*{-\algorithmicindent}\textbf{#1}%
\vspace*{-.7\baselineskip}\Statex\hspace*{\dimexpr-\algorithmicindent-2pt\relax}\rule{7.5cm}{0.4pt}%
}

\title{Learning Similarity between Movie Characters and Its Potential Implications on Understanding Human Experiences}
\author{Zhilin Wang\textsuperscript{1} $\;\;\;\;$ Weizhe Lin\textsuperscript{2} $\;\;\;\;$ Xiaodong Wu\textsuperscript{2} \\
  \textsuperscript{1}University of Washington, United States\\ \textsuperscript{2}University of Cambridge, United Kingdom\\
  \texttt{zhilinw@uw.edu, \{wl356, xw338\}@cam.ac.uk} \\}

\begin{document}
\maketitle
% \begin{abstract}
% This document is a supplement to the general instructions for *ACL authors. It contains instructions for using the \LaTeX{} style files for ACL conferences. 
% The document itself conforms to its own specifications, and is therefore an example of what your manuscript should look like.
% These instructions should be used both for papers submitted for review and for final versions of accepted papers.
% \end{abstract}

\input{main_content}

\bibliography{anthology,custom}
\bibliographystyle{acl_natbib}

\appendix
\section{Appendix A}
\label{sec:appendix}
\begin{table*}[b]
    \centering
    \begin{tabular}{lp{13cm}}
    \iffalse
        \midrule
        Reddit & I was shot and hit 3 times. I can’t talk to anyone in person about it. \\
        \midrule
        Movie characters & 1. Played for Laughs in a scene in horror movie parody Scary Movie: He then just sits and drinks his soda while all the other audience members continue to stab her to death. \\
        & 2. Smosh has a parody called "Pokémon In Real Life" starring Anthony: Anthony then finds a brick and yells, "TAKE THIS!!!" Then, he throws the brick at said bush and it hits Jigglypuff in the head\\
        & 3. Darkwing Duck's two-parter "Just Us Justice Ducks" ends with Negaduck falling off of city hall, bouncing off of power lines, having his head smashed against heavier and heavier objects to knock him back down \\
        & 4. Team Four Star's Let's Play of Left 4 Dead 2 custom campaign I Hate Mountains.: At one point Takahata has been incapped and Kaiser Neko asks if anyone wants to help him. Lani just says he's shooting Taka \\
        & 5. By the time Muttley finished rebuilding it, the land owner learned Dastardly tricked him to stop the other shortcut users and started shooting him. \\
    \fi
    \midrule
    Reddit post & My father passed away when I was 6 so I didn’t really remember much of him but the fact that I didn’t recognize his picture saddens me. \\
    \midrule
    Movie characters & 1. \textbf{Sisko in Star Trek: Deep Space Nine (Past Tense)} When he encountered an entry about the historical figure, passed comment about how closely Sisko resembled a picture of him (the picture, of course, being that of Sisko.)\\
    & 2. \textbf{Arator the Redeemer in World of Warcraft} As Arator never knew his father, he asks several of the veteran members of Alliance Expedition about Turalyon for information and leads on Turalyon's current location. Several people then gave their opinion on how great a guy Turalyon was, but sadly, he has been MIA for 15 years. \\
    & 3. \textbf{Kira in Push} The reality of a photo taken at Coney Island is the key evidence that causes her to realize that this was a fake memory. \\
    & 4. \textbf{Todd Aldridge in Mindwarp}  Todd shows up back in town; to him, there was a bright light one night, and he returned several months later with no knowledge of the intervening period. \\
    & 5. \textbf{Parker Girls in Stranger in Paradise} However, when the operation collapsed after the death of Darcy Parker many Parker Girls were trapped in their cover identities, unable to extricate themselves from the lives they had established. \\
    
    \midrule
    Reddit post & The black ladies I work with make me feel the most loved I've felt in years. I've had a horrible past 10 years. Childhood trauma and depression, addiction, abuse etc \\
    \midrule
    Movie characters & 1. \textbf{Shinjiro Aragaki in Persona 3} First of all, he's an orphan. During those two years, he began taking drugs to help control his Persona. Said drugs are slowly killing him. He has his own Social Link with the female protagonist where it becomes painfully clear that he really is a nice guy, and he slowly falls in love with her.\\
    & 2. \textbf{Mami in Breath of Fire IV} Country Mouse finds King in the Mountain God-Emperor that The Empire (that aforementioned God-Emperor founded) is trying very, very hard to kill. Country Mouse Mami nurses God-Emperor Fou-lu back to health. Mami and Fou-lu end up falling in love. \\
    & 3. \textbf{Emi in Katawa Shoujo} The loss of her legs was traumatic, but she learned to cope with that well. The loss of her dad she did not cope with at all. Part of getting her happy ending is to help her deal with her loss. \\
    & 4. \textbf{Harry in Harry Potter} Harry reaches out, has friends, and even in the moments when the school turns against him, he still has a full blown group of True Companions to help him, thus making him well adjusted and pretty close to normal. \\
    & 5. \textbf{Commander Shepard in the Mass Effect series} If the right dialogue is chosen, s/he's cynical and bitter with major emotional scars from his/her past experiences. It becomes pretty clear how emotionally burned out s/he really is.\\
    \end{tabular}
    \caption{Excepts from Posts from Reddit r/OffMyChest to five similar movie characters. Excerpts of Reddit posts mildly paraphrased to protect anonymity.}
    \label{tab:reddit_to_movie_characters}
\end{table*}

\end{document}

%% file: main_content.tex
%We conducted a pioneering study on finding most similar movie characters through a data-set of concise character descriptions. Specifically, we invented a simple but useful method of using the BERT Next Sentence Prediction (NSP) model to find most similar movie characters, leading to more than 250\% the performance of methods employing state-of-the-art paragraph embedding. Our method involves identifying a tiny proportion of all movie character-pairs to be compared rather than comparing all character-pairs. This tiny proportion includes only text-pairs on text-pairs whose BERT Masked Language Model average embedding are most cosine similar to each other, allowing the runtime complexity of this procedure to be reduced from $O(n^2)$ to $O(n)$ with respect to the number of characters. This represents a first step into understanding similarities among human characteristics and experiences through exploiting such parallels between movie characters and humans. 

\begin{abstract}
  While many different aspects of human experiences have been studied by the NLP community, none has captured its full richness. We propose a new task\footnote{Code and data available at \url{https://github.com/Zhilin123/similar_movie_characters}} to capture this richness based on an unlikely setting: movie characters. We sought to capture theme-level similarities between movie characters that were community-curated into 20,000 themes. By introducing a two-step approach that balances performance and efficiency, we managed to achieve 9-27\% improvement over recent paragraph-embedding based methods. Finally, we demonstrate how the thematic information learnt from movie characters can potentially be used to understand themes in the experience of people, as indicated on Reddit posts.
\end{abstract}

\section{Introduction}

What makes a person similar to another? While there is no definitive answer, some aspects that have been investigated in the NLP community are personality\citep{gjurkovic-snajder-2018-reddit, CONWAY201677}, demographics\citep{dong2016} as well as personal beliefs and intents
\citep{Sap_Le_Bras_Allaway_Bhagavatula_Lourie_Rashkin_Roof_Smith_Choi_2019}. While each of these aspects is valuable on its own, they also seem somewhat lacking to sketch a complete picture of a person. Researchers who recognise such limitations seek to ameliorate them by jointly modelling multiple aspects at the same time \citep{benton-etal-2017-multitask}. Yet, we intuitively know that as humans, we are more than the sum of the multiple aspects that constitutes our individuality. Our human experiences are marked by so many different aspects that interact in ways that we can not anticipate. What then can we do to better capture the degree of similarity between different people?

Finding similar movie characters can be an interesting first step to understanding humans better. Many characters are inspired by and related to true stories of people so understanding how to identify similarities between character descriptions might ultimately help us to better understand similarities in human characteristics and experiences. One way of defining what makes movie character descriptions similar is when community-based contributors on All The Tropes\footnote{\url{https://allthetropes.org}} classify them into the same theme (also known as a trope), with an example from the trope ``Driven by Envy'' shown in Table \ref{tab:charactor_description}. Other themes (tropes) include ``Parental Neglect'', ``Fallen Hero'', and ``A Friend in Need''. 

Such community-based curation allows All The Tropes to reap the same advantages as Wikipedia and open-sourced software: a large catalog can be created with high internal-consistency given the in-built self-correction mechanisms. This approach allowed us to collect a dataset of $>$100 thousand characters labelled with $>$20,000 themes without requiring any annotation cost. Based on this dataset, we propose a model that can be used to identify similar movie characters precisely yet efficiently. While movie characters may not be the perfect reflection of human experience, we ultimately show that they are good enough proxies when collecting a dataset of similar scale with real people would be extremely expensive. 

%Our work presents a rich and diverse dataset containing characters labelled with $>$125 thousand themes $>$20 thousand themes. Because collecting data of a similar scale from real people would be extremely difficult/expensive, our work presents a unique opportunity to understand what characters in common themes share. 

%For instance, real-life behaviors during COVID-19 might be understandable through characters in Steven Soderbergh's Contagion or Wolfgang Petersen's Outbreak. 

%We created a task using highly concise character descriptions on various media (movies/novels/anime) from All The Tropes\footnote{\url{https://allthetropes.org}}.We created Ground-Truth positive labels of closest neighbors based on whether these character descriptions share a common theme (also known as a trope), with an example from the trope ``Driven by Envy'' shown in Table \ref{tab:charactor_description}. Other themes (tropes) include ``Parental Neglect'', ``Fallen Hero'', and ``A Friend in Need''.
%Not only is the task suitable for testing our method of identifying text-pairs, but we also found the task intrinsically motivating.
%This is because many of these character descriptions are inspired by and related to true stories of people so understanding how to identify similarities between character descriptions might ultimately help us to better understand similarities in human characteristics and experiences as well. 

\begin{table}[!h]
\begin{center}
\begin{tabular}{|p{\linewidth}|}
\textbf{Superman's 1990s enemy Conduit.}

%This is also a facet of both the Master Jailer's hatred of Superman and that of Lex Luthor himself.
Conduit hates Superman because he knows if Superman wasn't around he would be humanity's greatest hero instead ... \\
%and he 
%he's the only person alive who is simply better than him,
%In Blackest Night, Wonder Woman used her lasso to make Lex admit that he wanted to be Superman. \\
\vspace{0.1cm}
\textbf{Loki}

%One of the reasons Loki's relationship with his brother Thor was so conflicted was because Loki envied Thor for being Odin's favored son.
Loki's constant scheming against Thor in his efforts to one-up him gave Odin and the rest of Asgard more and more reasons to hate Loki ...

%to the point that now every Asgardian except Thor wants to kill Loki even after he was reincarnated as a relatively innocent and powerless child with no memories of his past self.\\
\end{tabular}
\end{center}
\caption{Character descriptions from the trope ``Driven by Envy''}
\label{tab:charactor_description}
\end{table}

Our key contributions are as follows:
\begin{enumerate}
    \item We conduct a pioneering study on identifying similar movie character descriptions using weakly supervised learning, with potential implications on understanding similarities in human characteristics and experiences.
    
    \item We propose a two-step generalizable approach that can be used to identify similar movie characters precisely yet efficiently and demonstrate that our approach performs at least 9-27\% better than methods employing recent paragraph embedding-based approaches. 
    %This makes it feasible to use the BERT Next Sentence Prediction (NSP) model architecture, which is precise but would otherwise have too high a computational complexity to be used. 
    %\lin{We verified the rationale of this approach, and it could possibly enable all the research work that requires pair-wise comparison with a time-consuming model.}
    
    \item We show that our model, which is trained on identifying similar movie characters, can be related to themes in human experience found in Reddit posts. 
    %This shows the benefit of including pairwise comparisons (such as BERT NSP) in finding similar characters.
    
    %the necessity of adopting a pair-wise approach to find similar entries in a large dataset.
    
    %identifying a tiny fraction of text-pairs on which 
    %has to be performed 

%  \item The discovery of a simple but effective method of finding a tiny proportion of text-pairs on which to perform BERT Next Sentence Prediction, while maintaining good performance. 
%   Our method identifies text-pairs whose BERT Masked Language Modelling average embedding is most cosine similar to each other.
%  \item Doing so makes it feasible to use BERT Next Sentence Prediction to find paragraph-level text that is most similar to one another in a corpus of more than thousands of text (when it is previously not).
%  \item To evaluate the method and compare it with other existing models, we create a novel task involving finding character descriptions that share a common theme (also known as a trope).
\end{enumerate}

\section{Related Work}

\subsection{Analysis of characters in film and fiction}

Characters in movies and novels have been computationally analyzed by many researchers.
\citet{bamman2013learning, bamman2014bayesian} attempted to cluster various characters into prototypes based on topic modelling techniques \citep{Blei:2003:LDA:944919.944937}. On the other hand, \citet{frermann2017inducing} and \citet{iyyer2016feuding} sought to cluster fictional characters alongside the relationships between them using recurrent neural networks and matrix factorization. While preceded by prior literature, our work is novel in framing character analysis as a supervised learning problem rather than an unsupervised learning problem. 

Specifically, we formulate it as a similarity learning task between characters. Tapping on fan-curated movie-character labels (ie tropes) can provide valuable information concerning character similarity, which previous literature did not use. A perceptible effect of this change in task formulation is that our formulation allows movie characters to be finely distinguished amongst $>$ 20000 themes versus $<$ 200 in prior literature. Such differences in task formulation can contribute a fresh perspective into this research area and inspire subsequent research. 

Furthermore, the corpus we use differs significantly from those used in existing research. We use highly concise character descriptions of around 200 words whereas existing research mostly uses movie/book-length character mentions. Concise character descriptions can exemplify specific traits/experiences of characters. This allows the differences between characters to be more discriminative compared to a longer description, which might include more points of commonality (going to school/work, eating and having a polite conversation). This means that such concise descriptions can eventually prove more helpful in understanding characteristics and experiences of humans.

\subsection{Congruence between themes in real-life experiences and movie tropes}

Mostly researched in the field of psychology, real-life experiences are often analyzed through asking individuals to document and reflect upon their experiences. Trained analysts then seek to classify such writing into predefined categories. 

\citet{demorest1999comparison} interpreted an individual’s experience in the form of three key stages: an individual’s wish, the response from the other and the response from the self in light of the response from the other. Each stage consists of around ten predefined categories such as wanting to be autonomous (Stage 1), being denied of that autonomy (Stage 2) and developing an enmity against the other (Stage 3). 
\citet{thorne2001} organized their analysis in terms of central themes. These central themes include experiences of interpersonal turmoil, having a sense of achievement and surviving a potentially life-threatening event/illness.

Both studies above code individuals' personal experiences into categories/themes that greatly resemble movie tropes. 
Because of this congruence, it is very likely that identifying similarity between characters in the same trope can inform about similarity between people in real-life. A common drawback of \citet{demorest1999comparison} and \citet{thorne2001} lie in their relatively small sample size (less than 200 people classified into tens of themes/categories). Comparatively, our study uses $>$ 100,000 characters fine-grainedly labelled by fans into $>$20,000 tropes. As a result, this study has the potential of supporting a better understanding of tropes, which we have shown to be structurally similar to themes in real-life experiences. 

\subsection{Candidate selection in information retrieval}

Many information retrieval pipelines involve first identifying likely candidates and then post-processing these candidates to determine which among them are most suitable. The most widely-used class of approaches for this purpose is known as Shingling and Locally Sensitive Hashing \citep{leskovec_rajaraman_ullman_2020, rodier-carter-2020-online}. Such approaches first represent documents as Bag-of-Ngrams before hashing such representation into shorter integer-vector signatures. These signatures contain information on n-gram overlap between documents and hence encode lexical features that characterize similar documents. However, such approaches are unable to identify documents that are similar based on abstract semantic features rather than superficial lexical similarities. 

Recent progress in language modelling has enabled the semantic meaning of short paragraphs to be encoded beyond lexical features \citep{peters2018deep, devlin-etal-2019-bert, howard2018universal,Raffel2019ExploringTL}. This has reaped substantial gains in text similarity tasks including entailment tasks \citep{snli:emnlp2015, N18-1101}, duplicate questions tasks \citep{DBLP:journals/corr/abs-1907-01041,nakov-etal-2017-semeval} and others \citep{cer-etal-2017-semeval, dolan-brockett-2005-automatically}.
Yet, such progress has yet to enable better candidate selection based on semantic similarities. As a result, relatively naive approaches such as exhaustive pairwise comparisons and distance-based measures continue to be the dominant approach in identifying similar documents encoded into dense contextualized embeddings \citep{reimers-gurevych-2019-sentence}. To improve this gap in knowledge, this study proposes and validates a candidate selection method that is compatible with recent progress in text representation.

\section{Task formulation}

There is a set of unique character descriptions from the All The Tropes ($Character_0$, $Character_1$ ... $Character_n$), each associated with a non-unique trope (theme) ($Trope_0$, $Trope_0$ ... $Trope_p$). Given this set, find the $k$ (where $k$ = 1, 5 or 10) most similar character(s) to each character without making explicit use of the trope association of each character. In doing so, the goal is to have a maximal proportion of most similar character(s) which share the same tropes. 

\section{Methods}

\begin{figure*}
    \centering
    \includegraphics[width=12cm]{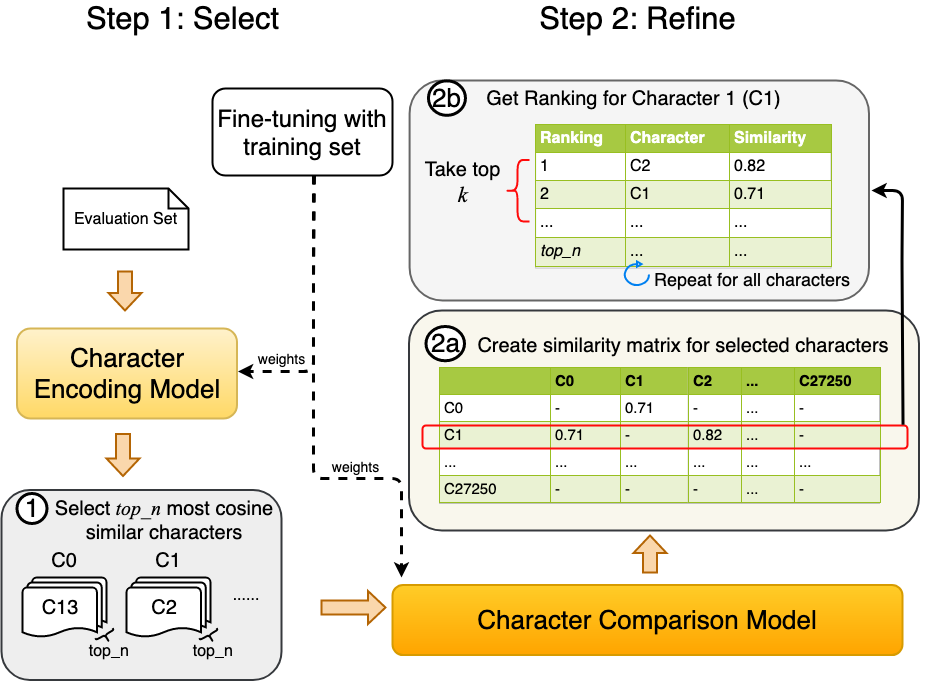}
    \caption{Workflow of finding most similar characters: BERT NSP model is first trained on the training set (Section \ref{sec:fine-tuning_model}) $top\_n$ characters are then selected using cosine similarity based on the Character Embedding Model or using a Siamese-BERT model, which has been omitted from the illustration for clarity (Section \ref{sec:select}).
    This selection is then refined using the Character Comparison Model to create a similarity matrix, which can then be sorted to identified most similar characters. 
    (Section \ref{sec:refine})
    }
    \label{fig:workflow}
\end{figure*}

In this section, we first discuss how we prepare the dataset and trained a BERT Next Sentence Prediction (NSP) model to identify similar characters. 
%Then, we test the feasibility of using this fine-tuned model to identify similar characters. 
Based on this model, we present a 2-step \textbf{Select} and \textbf{Refine} approach, which can be utilized to find the most similar characters quickly yet effectively. 

%of finding the closest neighbors. 
% The workflow is presented in Fig. \ref{fig:workflow}

\subsection{Dataset}\label{sec:dataset_preparation}

Character descriptions from All The Tropes\footnote{\url{https://allthetropes.org}} were used. We downloaded all character descriptions that had more than 100 words because character descriptions that are too short are unlikely to provide sufficient textual information for comparing similarity with other character descriptions. We then filtered our data to retain only tropes that contain more than one character descriptions.  Character descriptions were then randomly split into training and evaluation sets (evaluation set = 20\%). Inspired by BERT NSP dataset construction \citet{devlin-etal-2019-bert}, we generated all possible combination-pairs of character descriptions that are classified under each trope (i.e. an unordered set) and gave the text-pair a label of \texttt{IsSimilar}, For each \texttt{IsSimilar} pair in the training set, we took the first item, randomly selected a character description that is not in the same trope as the first item and gave the new pair a label of \texttt{NotSimilar}. 

%This resulted in around 300,000 character descriptions that are classified under 22,000 tropes (themes).
%As a result, this gave us a balanced training set of 3.5 million text-pairs (approximately 7 GBs).

Descriptive statistics are available in Table \ref{tab:descriptive_statistics}.

\begin{table*}[!h]
\centering
\begin{tabular}{lll}
\toprule
& Training Set & Evaluation Set\\
\midrule
Characters   & 109000         & 27250 \\ % 269586         & 29954
Words per character   & 172 ($\sigma$ = 101)         &  172 ($\sigma$ = 102)\\
Tropes   & 13160         & 8669 \\
Characters per trope  & 5.39 ($\sigma$ = 9.66)         & 1.33 ($\sigma$ = 2.64)\\
Character-pairs & 2375298 (50\% \texttt{IsSimilar}) & 72705 (only \texttt{IsSimilar})  \\ %3123496 & 73992
\bottomrule
\end{tabular}
\caption{Descriptive statistics of dataset}
\label{tab:descriptive_statistics}
\vspace{-0.3cm}
\end{table*}

\subsection{Training BERT Next Sentence Prediction model}\label{sec:fine-tuning_model}

We trained a BERT Next Sentence Prediction model (English-base-uncased)\footnote{12-layer, 768-hidden, 12-heads, 110M parameters with only Next Sentence Prediction loss, accessed from \url{https://github.com/huggingface/transformers}} with the pre-trained weights used as an initialization.
As this model was trained to perform pair-wise character comparison instead of next sentence prediction, we thereafter name it as Character Comparison Model (\texttt{CCM}). % to prevent any confusion. Fine-tuning the model was necessary because our task contains language used in an informal context while the original BERT Next Sentence Prediction model was trained on the language used in a formal context (Wikipedia and Books).Furthermore, the model was originally pre-trained on a sentence-level task whereas our task is a paragraph-level one.

All hyper-parameters used to train the model were default\footnote{\url{https://github.com/huggingface/transformers/}} except adjusting the maximum sequence length to 512 tokens (to adapt to the paragraph-length text), batch-size per GPU to 8 and epoch number to 2, as recommended by \citet{devlin-etal-2019-bert}. Among the training set, 1\% was separated as a validation set during the training process.
We also used the default pre-trained BERT English-base-uncased tokenizer because only a small proportion of words ($<$ 0.5\%) in the training corpus were out-of-vocabulary, of which most were names.
%Therefore, it was not justifiable to use additional resources to re-train a tokenizer.
As a result, training took 3 days on 4 Nvidia Tesla P100 GPUs.

\subsection{Select and Refine}\label{sec:findclosestneighbors}

To address the key limitation of utilizing exhaustive pairwise comparison in practice - its impractically long computation time ($\approx$ 10 thousand GPU-hours on Nvidia Tesla P100), we propose a two-step Select and Refine approach. The Select step first identifies a small set of likely candidates in a coarse but computationally efficient manner. Then, the Refine step re-ranks these candidates using a precise but computationally expensive model. In doing so, it combines their strengths to precisely identify similar characters while being computationally efficient. While the Select and Refine approach is designed for identifying similar characters, this novel approach can also be directly used in other tasks involving semantic similarities between a pair of texts. 

%Exhaustive character-pair comparison was not used as a baseline because it could take for close to 30,000 characters in the evaluation set. 

\subsubsection{Select}\label{sec:select}

Characters that are likely to be similar to each character are first selected using a variant of our \texttt{CCM} model - named the Character Encoding Model (thereafter \texttt{CEM}). This model differs from the \texttt{CCM} model in that it does not utilize the final classifier layer. Therefore it can process a character description individually (instead of in pairs) to output an embedding that represents the character. The shared weights with \texttt{CCM} means that it encodes semantic information in a a similar way. This makes it likely that the most cosine similar character descriptions based on their character embedding are likely to have high (but not necessarily the highest) character-pair similarity. 

Beyond the \texttt{CEM}, any model capable of efficiently generating candidates for similar character description texts in O(n) time can also be used for this Select step, allowing immense flexibility in the application of the Select and Refine approach. To demonstrate this, we also test a Siamese-BERT model for the Select step, with the details of its preparation in Section \ref{sec:baselines}.

In this step, we effectively reduced the search space for the most similar characters. We choose $top\_n$ candidates characters which are most similar to each character, forming $top\_n$ most similar character-pairs. $top\_n$ is a hyper-parameter that can range from 1 to 500. %This procedure is outlined in Algorithm \ref{alg:most_similar}. 
Strictly speaking, this step requires $O(n^2)$ comparisons to find the $top\_n$ most similar character-pairs. However, each cosine similarity calculation is significantly less computationally demanding compared to each BERT NSP operation (note that \texttt{CCM} is trained from an NSP model). This also applies to the Siamese-BERT model because character embeddings can be cached, meaning that only a single classification layer operation needs to be repeated $O(n^2)$ times. This means that computational runtime is dominated by $O(n)$ BERT NSP operations in the subsequent Refine step, given the huge constant factor for BERT NSP operations. Overall, this step took 0.25 GPU-hours. 

% Values ranging from 10 to 500 were experimented with for top\_n.
%- as well as a few orders of magnitude larger than this number of characters.
%Argue that it's O(n) to get the embeddings for the Select step but $O(n^2)$ for the classifier for the similarity. Because character embeddings can be cached, this is O(n)

\subsubsection{Refine}\label{sec:refine}

The initial selection of candidates for most similar characters to each character will then be refined using the \texttt{CCM} model. This step is more computationally demanding (0.25 * $top\_n$ GPU-hours) but can more effectively determine the extent to which characters are similar. 
%In step \textbf{2a} of Figure \ref{fig:workflow} (corresponding to Algorithm \ref{alg:matrix}a),
Character Comparison Model (\texttt{CCM}) will then only be used on the $top\_n$ most similar candidate character-pairs, reducing the number of operations for each character from the total number of characters ($n_{chars}$) to only $top\_n$. As a consequence, the runtime complexity of the overall operation is reduced from $O(n_{chars}^2)$ to $O(top\_n \cdot n_{chars})$ == $O(n_{chars})$, given $top\_n$ is a constant.

\section{Evaluation}

In this section, we first present evaluation metrics and then present the preparation of baseline models including state-of-the-art paragraph-level embedding models. Finally, we analyze the performance of our models relative to baseline models. %Finally, we demonstrate the potential implications of our findings for understanding themes in real-life experiences. %Finally, we discuss how varying $top\_n$ - a critical hyper-parameter that controls the number of most cosine similar characters to be compared to each character - might affect the performance of our model . %(which controls the number of most similar text to be engaged in similarity matrix creation).

\subsection{Evaluation metrics}

\textbf{Recall @ k} considers the proportion of all ground-truth pairs found within the k (1, 5 or 10) most similar characters to each character \citep{manning2008introduction}. Normalized Discounted Cumulative Gain @ k (\textbf{nDCG @ k}) is a precision metric that considers the proportion of predicted k most similar characters to each character that are in the ground-truth character-pairs. It also takes into account the order amongst top k predicted most similar characters \citep{Wang2013ATA}. Mean reciprocal rank (\textbf{MRR}) identifies the rank of the first correctly predicted most similar character for each character and averages the reciprocal of their ranks. \citep{voorhees2000trec}. Higher is better for all metrics. 

\subsection{Baseline Models}\label{sec:baselines}

Baseline measurements were obtained for Google Universal Sentence Encoder-large \citep{cer2018universal}, BERT-base \citep{devlin-etal-2019-bert} and Siamese-BERT-base\footnote{12-layer, 768-hidden, 12-heads and 110M parameters} \citep{reimers-gurevych-2019-sentence}. 
%Word2Vec \citep{mikolov2013efficient}, and RoBERTa-base embeddings \citep{liu2019roberta}

Google Universal Sentence Encoder-large model\footnote{https://tfhub.dev/google/universal-sentence-encoder-large/3} (\textbf{USE}) on Tensorflow Hub was used to obtain a 512-dimensional vector representation of each character description. Bag of Words (\textbf{BoW}) was implemented by lowercasing all words and counting the number of times each word occurred in each character description. \textbf{BERT} embedding of 768 dimensions were obtained by average-pooling all the word embedding of tokens in the second-to-last layer, as recommended by \citep{xiao2018bertservice}. The English-base-uncased version\footnote{12-layer, 768-hidden, 12-heads and 110M parameters} was used. For each type of embedding, the most similar characters were obtained by finding other characters whose embeddings are most cosine similar. 

\textbf{Siamese-BERT} was obtained based on training a Siamese model architecture connected to a BERT base model on the training set in Section \ref{sec:dataset_preparation}. We follow the optimal model configuration for sentence-pair classification tasks described in \citet{reimers-gurevych-2019-sentence}, which involves taking the mean of all tokens embeddings in the final layer. With the mean embedding for each character description, an absolute difference between them was taken. The mean embedding for character A, mean embedding for character B and their absolute difference was then entered into a feedforward neural network, which makes the prediction. Siamese-BERT was chosen as a baseline due to its outstanding performance in sentence-pair classification tasks such as Semantic Textual Similarity \citep{cer-etal-2017-semeval} and Natural Language Inference \citep{snli:emnlp2015, N18-1101}. For this baseline, the characters most similar to a character are those with the highest likelihood of being predicted \texttt{IsSimilar} with the character. 

%BERT CLS embedding, also of 768 dimensions were obtained by extracting the embedding of the CLS token in the second to last layer. \textbf{RoBERTa} embedding of 768 dimensions was extracted using PyTorch Hub \citep{robertahub}. Text embedding for each character description was created by averaging all word embedding in the last layer, as recommended as a method to extract features for downstream tasks. The version used was the base model.\footnote{12-layer, 768-hidden, 12-heads and 110M parameters} \textbf{Word2Vec} was implemented by using pre-trained 300-dimensional Word2Vec embedding\footnote{https://code.google.com/archive/p/word2vec/}.Character descriptions were first tokenized, lemmatized and stripped of stop-words before being transformed into a Bag of Words corpus. Thereafter, the embedding for each character description was calculated by taking a simple mean of all word embedding.

\subsection{Suitability of Siamese-BERT and CEM for Step 1: Select}

While the prohibitively high computational demands of exhaustive pairwise comparison ($\approx$ 10 thousand GPU-hours) prevents a full-scale evaluation  of the adequateness of Siamese-BERT and \texttt{CEM} for Step 1:Select, we conducted a small-scale experiment on 100 randomly chosen characters from the test set. First, an exhaustive pairwise comparison was conducted between these randomly chosen characters and all characters in the test set. From this, 100 characters with the highest \texttt{CCM} similarity value with each of the randomly chosen characters were identified. Next, various methods in Table \ref{tab:proportion_overlap} were attempted to identify 500 characters with the highest cosine similarity with the randomly chosen characters. Finally, the proportion of overlap between \texttt{CCM} and each method was calculated. Results demonstrate that Siamese-BERT and \texttt{CEM} have the greatest overlap and hence, the use of Siamese-BERT and \texttt{CEM} can select for the most number of highly similar characters to be refined by the \texttt{CCM}. 

%We first identified 500 characters whose \texttt{CEM} embedding are most cosine similar to each of the 100 randomly chosen characters. These 500 most cosine similar characters were found to overlap with 24.2\% of the  . This proportion is higher than other methods that can be used in this step, as shown in Table \ref{tab:proportion_overlap}. Therefore,

\begin{table}[!h]
\centering
\begin{tabular}{ll}
\toprule
& \texttt{CCM} overlap (\%)\\
\midrule
Siamese-BERT & \textbf{36.15}\\ 
\texttt{CEM }& 24.21\\ 
BERT & 16.90\\
%BERT CLS & 9.90\\
USE & 16.27 \\
%RoBERTa average & 18.73\\
%Word2Vec average & 16.36\\ %3123496 & 73992
BoW & 7.41\\
\bottomrule
\end{tabular}
\caption{Proportion of 100 characters with high \texttt{CCM} similarity value that overlaps with each method for Step 1: Select}
\label{tab:proportion_overlap}
\vspace{-0.4cm}
\end{table}

\subsection{Selecting hyper-parameter top\_n for Step 2: Refine}\label{sec:hyperparameter}

\begin{figure}[h]
    \centering
    \includegraphics[width=6.5cm]{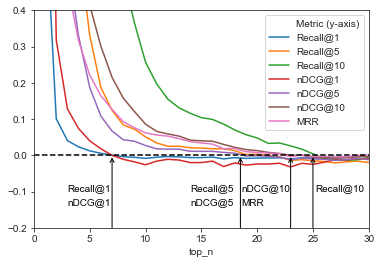}
    \caption{Percent change in metrics with each additional $top\_n$ for Select and Refine model with Siamese-BERT. Average smoothing applied over a range of 10 to improve clarity. Points annotated where each metric is at 0.}
    \label{fig:percent_change_per_additional_n1}
\end{figure}

Based on Figure \ref{fig:percent_change_per_additional_n1}, the ideal $top\_n$ for the Select and Refine model with Siamese-BERT varies between 7 and 25 depending on the metric that is optimised for. In general, a lower value for $top\_n$ is preferred when optimizing for Recall@k and nDCG@k with smaller values of k. The metrics reported in Table \ref{tab:updated_compared_to_baseline_measures} consist of the optimal value for each metric at various $top\_n$.

On the other hand, there is no ideal value for $top\_n$ when using the Select and Refine model with \texttt{CEM}. Instead, the metrics continue to improve over large values of $top\_n$, albeit at a gradually reduced rate. However, due to practical considerations relating to GPU computation time, we terminated our search at $top\_n$ = 500 and report metrics for that value of $top\_n$.

Together, this means that the Select and Refine model using Siamese-BERT achieves peak performance with significant less computational resources compared to the one using \texttt{CEM} (2-6 GPU-hours vs. 125 GPU-hours). 

\subsection{Comparing Select and Refine models with baseline models}

\begin{table*}[!htp]
\centering
\begin{tabular}{llllllll}
\toprule
& \multicolumn{3}{c|}{\textbf{Recall @ k }(in \%)} & \multicolumn{3}{c|}{\textbf{nDCG @ k }(in \%)} & \textbf{MRR} \\
& k = 1 & k = 5 & k = 10 & k = 1 & k = 5 & k = 10 & (in \%)\\
\midrule
\textbf{Select and Refine models}\\
%\midrule
%Select using \\
\midrule
%Initialize top 10         & \textbf{2.642}          & \textbf{5.939}           & 6.352 \\
%best performance for recall and ndcg @ 1 = 10; @5 + MRR --> 20; @10 --> 30
Siamese-BERT  & \textbf{6.921} & \textbf{23.53} & \textbf{36.14} &  \textbf{21.82} &\textbf{19.13} & \textbf{19.71} & 26.56 \\
\texttt{CEM} & 6.184 &  20.74 &  31.02 &  19.50 &  17.02 & 17.26 &  \textbf{28.50}\\ %24.83 * 27439/23901
%CEM &  100 & 5.704 &  18.35 &  26.81 &  17.98 &  15.15 & 15.11  &  22.75*27439/23901 \\
\midrule
\textbf{Baseline models} \\
\midrule
%& Finetuned?\\
%\midrule
Siamese-BERT  &5.437 &  19.53 &  30.65 &  17.14 &  15.51 &  16.15 &  25.99 \\

\texttt{CEM}  &2.802 &  8.852 &  13.26 &  8.832 &  7.119 & 7.126  &  14.61  \\ %12.73 *27439/23901
%BERT fine-tuned CLS &  &3.218 &  10.88 &  16.78 &  10.15 &  8.412 & 8.422 & 14.09\\
%this was previously BERT average
BERT          &1.238 &  3.636 &  5.514 &  3.904 &  3.035 & 3.109  & 7.182 \\ %6.256*27439/23901
USE           &1.277 &  4.427 &  6.956 &  4.025 &  3.599 &  3.866 & 7.810  \\ %6.803*27439/23901
BoW           &0.4632 &  1.344 &  1.987 &  1.46 &  1.087 &  1.052 & 2.824 \\ %2.46*27439/23901
\bottomrule

\end{tabular}
\caption{Performance of Select and Refine models compared to baseline models. Higher is better for all metrics.}
\label{tab:updated_compared_to_baseline_measures}
\end{table*}

\iffalse
    Ours is actually a similarity learning task so we can have more similarity learning baselines?
    
    This is on social media
    https://ieeexplore.ieee.org/abstract/document/8683405

    Note they are all in CV
    
    Such as https://openaccess.thecvf.com/content_cvpr_2018/papers/Pinheiro_Unsupervised_Domain_Adaptation_CVPR_2018_paper.pdf
    
    Or https://papers.nips.cc/paper/8747-neural-similarity-learning.pdf
    
    or Large Scale Similarity Learning Using Similar Pairs for Person Verification
    
    or https://openaccess.thecvf.com/content_CVPR_2019/papers/Wang_Multi-Similarity_Loss_With_General_Pair_Weighting_for_Deep_Metric_Learning_CVPR_2019_paper.pdf
\fi

As shown in Table \ref{tab:updated_compared_to_baseline_measures}, the highest value for all metrics lies below 40\% suggesting that identifying similar characters is a novel and challenging task. This is because there are only very few correct answers (characters from the same trope) out of 27,000 possible characters. The poor performance of the Bag-of-Words baseline also demonstrates that abstract semantic similarity between characters is significantly different from their superficial lexical similarity. In face of such challenges, the Select and Refine model using Siamese-BERT performed 9-27 \% better on all metrics than the best performing paragraph-embedding-based baseline. This suggests the importance of refining initial selection of candidates instead of using them directly, even when the baseline model has relatively good performance. 

Comparing the Select and Refine models, Siamese-BERT performed much better than \texttt{CEM} while having a significantly low $top\_n$, which means that less computational resources is required. The superior performance and efficiency of Siamese-BERT means that it is more suitable for Step 1: Select. This is likely caused by the higher performance of Siamese-BERT as a baseline model. While it was surprising that using Siamese-BERT outperformed \texttt{CEM}, which directly shares weights with the \texttt{CCM}, such an observation also shows the relatively low coupling between the Select and Refine steps. This means that the Select and Refine approach that we propose can continue to be relevant when model architectures that are more optimized for each step are introduced in the future. 

The significantly higher performance of Select and Refine models can be attributed to the ability of underlying BERT NSP architecture in our \texttt{CCM} to consider complex word relationships across the two character descriptions. A manual examination of correct pairs captured only by Select and Refine models but not baseline models revealed that these pairs often contain words relating to multiple common aspects. As an example, one character description contains “magic, enchanter” and “training, candidate, learn” while the other character in the ground-truth pair contains “spell, wonder, sphere” and “researched, school”. Compressing these word-level aspects into a fixed-length vector would cause some important semantic information - such as the inter-relatedness between aspects - to be lost \citep{conneau-etal-2018-cram}. As a result, capturing similarities between these pairs prove to be difficult in baseline models, leading to sub-optimal ranking of the most similar characters. 
%Considering correctly identified most similar character descriptions identified by our $top\_n$ = 500 model and the best baseline model of USE, 12\% of character-pairs were found in common, 22\% by USE only and 66\% by our model only. 

\section{Implications for understanding themes in real-life experiences}\label{sec:understand_real_life_experiences}

\subsection{Relating movie characters to Reddit posts}\label{sec:reddit_relation}

\begin{table*}[!htp]
\centering
\begin{tabular}{llll}
\toprule
& \multicolumn{3}{c}{\textbf{Precision @ k }(in \%)} \\
& k = 1 & k = 5 & k = 10 \\
\midrule
\textbf{Select and Refine models}\\

\midrule
Siamese-BERT  & \textbf{98.0} (14.0) & \textbf{92.4} (14.4) & \textbf{87.0} (8.79) \\
\texttt{CEM} & 82.0 (39.6) & 77.6 (17.9)  & 70.2 (8.94)\\ 
\midrule
\textbf{Baseline models} \\
\midrule
Siamese-BERT  & 76.0 (42.8) & 73.2 (14.9)  & 70.8 (8.31)\\
\texttt{CEM}  & 48.0 (38.4) & 33.2 (20.5)  & 27.2 (11.3) \\ 
BERT          & 40.0 (48.9) & 21.2 (12.7)  & 12.8 (5.98) \\ 
USE           & 32.0 (46.6) & 15.6 (13.3)  & 9.2 (7.23)\\ 
BoW           & 16.0 (36.6) & 7.2 (9.17)  & 4.4 (4.9) \\ 
\bottomrule

%\bottomrule
\end{tabular}
\caption{Precision @ k (std. dev.) for movie characters identified by each model.}
\label{tab:person_annotation}
%\vspace{-0.4cm}
\end{table*}

%Our work presents a rich and diverse dataset containing characters labelled with $>$20 thousand themes. Because collecting data of a similar scale from real people would be extremely difficult/expensive, our work presents a unique opportunity to understand what characters in common themes share. Thereafter, such themes can help to gain insights into the themes of real-life experiences through zero/few-shot transfer learning.  
To demonstrate the potential applications of this study in understanding human experiences, we designed a task that can show how the model can be used with zero-shot transfer learning. Specifically, we used our model to identify the movie-characters that are most fitting to a description of people's life experiences. To do this, we collected 50 posts describing people's real-life experiences from a forum r/OffMyChest on Reddit\footnote{\url{https://www.reddit.com/r/offmychest/}}, on which people share their life experiences with strangers online.

Then, we used our models to identify 10 movie characters (from our test set) that are most befitting to each post. For each of these 10 movie characters suggested by model, three graduate students independently rated whether the character matches the concepts, ideas and themes expressed in each post, while blind to information on which model the characters were generated by. Because the extent of similarity between a movie character and a Reddit post can be ambiguous, a binary annotation was chosen over a Likert scale for clarity of annotation. Annotators were instructed to annotate "similar" when they can \textbf{specify} at least one area of overlap between the concepts, ideas and themes of a Reddit post and a movie character. Examples of some characters that are indicated as "similar" to two posts are shown in Appendix  \ref{sec:appendix}. 
Annotators agree on 94.2\% of labels (Cohen's $\kappa$ = 0.934). Where the annotators disagree, the majority opinion out of three is taken. From these annotations, \textbf{Precision @ k} is calculated, considering the proportion of all characters identified within the k (1, 5 or 10) that are labelled as "similar" \citep{manning2008introduction}.

\begin{table*}[!ht]
    \centering
    \begin{tabular}{lp{13cm}}
    %\begin{adjustbox}{max width=\textwidth}
    \midrule
    \textbf{Reddit post} & My father passed away when I was 6 so I didn’t really remember much of him but the fact that I didn’t recognize his picture saddens me. \\
    \midrule
    \textbf{Select and Refine} \\
    \midrule
    Siamese-BERT & \textbf{Sisko in Star Trek: Deep Space Nine (Past Tense)} When he encountered an entry about the historical figure, passed comment about how closely Sisko resembled a picture of him (the picture, of course, being that of Sisko.)\\
    \texttt{CEM}  & \textbf{Roxas in Kingdom Hearts: Chain of Memories} His memories are wiped by Ansem the Wise and placed in a simulated world with a completely new identity \\
    \midrule
    \textbf{Baseline} \\
    \midrule
    Siamese-BERT &  \textbf{Audrina, My Sweet Audrina by V.C Andrews }is a girl living in the constant shadow of her elder sister who had died nine years before she was born \\
    \texttt{CEM} & \textbf{Macsen Wledig in The Mabinogion}  An amazing memory was an important necessity to the job, but remembering many long stories was much more important than getting one right after days of wandering around madly muttering \\
    BERT & \textbf{Kira in Push} is made to think that her entire relationship with Nick was a false memory that she gave him and she's been pushing his thoughts the entire time they were together.\\
    USE & \textbf{EyeRobot in Fallout: New Vegas} can recognize your face and voice with advanced facial and auditory recognition technology\\
    BoW & \textbf{Magneto} took Ron the Death Eater Up to Eleven to show him as he "truly" was in Morrison's eyes, and ended with him (intended as) Killed Off for Real\\
    \bottomrule
    %\end{adjustbox}
    \end{tabular}
    \caption{Most similar character predicted by each model to a post from Reddit r/OffMyChest. Excerpts of Reddit post mildly paraphrased to protect anonymity.}
    \label{tab:reddit_to_all_models_movie_characters}
\end{table*}

\iffalse
    \begin{table}[!h]
    \centering
    \begin{tabular}{ll}
    \toprule
    & \textbf{Precision@10} \\
    & (in \%) \\
    \midrule
    \textbf{Select-and-Refine models} \\
    \midrule
    Siamese-BERT & \textbf{36.15}\\ 
    \texttt{CEM}& 24.21\\ 
    \midrule
    \textbf{Baseline models} \\
    \midrule
    Siamese-BERT & 36.15\\ 
    \texttt{CEM}& 24.21\\ 
    BERT & 16.90\\
    USE & 16.27 \\
    BoW & 7.41\\
    %\bottomrule
    \end{tabular}
    \caption{Precision @ 10 for movie characters identified by each model.}
    \label{tab:person_annotation}
    %\vspace{-0.4cm}
    \end{table}
\fi 
%For instance, descriptions of people’s life experiences (e.g. social media posts/diary entries) might be automatically labelled as fitting a certain movie-character. With $>$ 100 thousand characters in $>$ 20 thousand themes, this approach can label people’s description of their life experiences in an ultra-fine-grained manner like never-before. To provide an illustration of how our model can be used zero-shot, we used our best model to find movie characters most similar to the experiences of people in real-life in Table  \ref{tab:reddit_to_movie_characters}. 

In Table \ref{tab:person_annotation}, the performance of our Select and Refine models reflects a similar extent of improvement compared to our main learning task. This shows that the model that was trained to disambiguate movie character similarity can also determine the extent of similarity between movie characters and people's life experiences. Beyond the relative performance gains, the Select and Refine model on this task also demonstrates an excellent absolute performance of precision @ 1 = 98.00\%. This means that our model can be used on this task without any fine-tuning. 

Illustrating the difference in performance of the various models in Table \ref{tab:reddit_to_all_models_movie_characters}, the better performing models on this task are generally better at capturing thematic similarities in terms of the abstract sense of recollection and memory, which are thematically more related to the Reddit post. Our Select and Refine model (with Siamese-BERT) is particularly effective at capturing both a sense of recollection as well as a sense of reverence towards a respected figure (historical figure and father respectively). In contrary, the poorer performing models contain phrase-level semantic overlap (USE: picture with facial recognition; BoW: killed and passed away; eyes and recognize) but fail to capture thematic resemblance. This suggests our learning of similarities between movie characters of the same trope can effectively transfer onto thematic similarities between written human experiences and movie characters. 

\subsection{Future directions}\label{sec:future}

We are excited about the diversity of research directions that this study can complement. One possible area is social media analysis \citep{zirikly-etal-2019-clpsych, amir-etal-2019-mental, hauser-etal-2019-using}. Researchers can make use of movie characters with known experiences (e.g. mental health, personal circumstances or individual interests) to identify similar experiences in social media when collecting large amounts of text labelled with such experiences directly is difficult. 

Another area would be personalizing dialogue agents \citep{tigunova-etal-2020-charm, zhang-etal-2018-personalizing}. In the context of limited personality-related training data, movie characters with personality that are similar to a desired dialogue agent can be found. Using this, a dialogue agent can be trained with movie subtitle language data (involving the identified movie character). Thereby, the augmented linguistic data enables the dialogue agent to have a well-defined, distinct and consistent personality. 

A final area that can benefit from this study is media recommendations \citep{RAFAILIDIS201711}. Users might be suggested media content based on the extent to which movie characters resonate with their own/friends' experiences. Additionally, with social environments being formed in games (particularly social simulation games such as Animal Crossing, The Sims and Pokemon) as well as in virtual reality \citep{hang_vr_2020}, participants can even assume the identity of movie characters that they are similar to, so as to have an interesting and immersive experience.

\section{Conclusion}

%\textbf{This means that such concise descriptions can eventually prove more helpful in understanding characteristics and experiences of humans. Conducted a pioneering study on finding most similar concise character descriptions, with potential implications on understanding similarities in human characteristics and experiences. Invented an easy but effective method of using BERT Next Sentence Prediction (NSP) in find most similar movie characters, leading to more than 250\% the performance of methods employing state-of-the-art paragraph embedding.}

We introduce a pioneering study on identifying similar movie characters through weakly supervised learning. Based on this task, we introduce a novel Select-and-Refine approach that allows us to match characters belonging to a common theme, which simultaneously optimize for efficiency and performance. Using this trained model, we demonstrate the potential applications of this study in identifying movie characters that are similar to human experiences as presented in Reddit posts, without any fine-tuning. This represents an early step into understanding the complexity and richness of our human experience, which is not only interesting in itself but can also complement research in social media analysis, personalizing dialogue agents and media recommendations/interactions. 

\section*{Acknowledgements}

We would like to thank the anonymous reviewers for their helpful feedback.

\newpage